\documentclass[11pt]{article}
\usepackage[utf8]{inputenc}
\usepackage[T1]{fontenc}
\usepackage[main=american]{babel}
\usepackage[babel=true]{csquotes}
\usepackage{natbib}
\usepackage{eamt20}
\usepackage{times}
\usepackage{url}
\usepackage{latexsym}
\usepackage{multirow}
\usepackage[table,xcdraw]{xcolor}
\usepackage{xspace}
\usepackage{tabularx}
\usepackage{booktabs}
\usepackage[small,bf]{caption} 
\usepackage{subcaption}
\usepackage{changepage}
\newcommand{\ie}{i.e.\xspace}

\newcommand{\eg}{e.g.\xspace}

\newcommand{\percent}{\%\,}

\newcommand{\citeg}[1]{\citeauthor{#1}'s \citeyearpar{#1}} 

\newcommand{\Section}[1]{Section~\ref{sec:#1}}
\newcommand{\Table}[1]{Table~\ref{tab:#1}}
\newcommand{\Figure}[1]{Figure~\ref{fig:#1}}

\title{What's the Difference Between Professional Human and Machine Translation? A Blind Multi-language Study on Domain-specific MT}

\author{Lukas Fischer\,$^{1}$ 
    \quad Samuel Läubli\,$^{1,2}$ \bigskip \medskip\\
    $^1$\,TextShuttle AG \medskip\\
    $^2$\,Department of Computational Linguistics, University of Zurich}

\date{}

\begin{document}
\maketitle
\begin{abstract}
    Machine translation (MT) has been shown to produce a number of errors that require human post-editing, but the extent to which professional human translation (HT) contains such errors has not yet been compared to MT. We compile pre-translated documents in which MT and HT are interleaved, and ask professional translators to flag errors and post-edit these documents in a blind evaluation. We find that the post-editing effort for MT segments is only higher in two out of three language pairs, and that the number of segments with wrong terminology, omissions, and typographical problems is similar in HT.
\end{abstract}

\section{Introduction}
\label{sec:Introduction}

Machine translation (MT) quality has improved substantially over the past years, allegedly to the degree that it is no longer distinguishable from professional human translation (HT). The first claims of human--machine parity were based on MT systems geared to news translation \citep{Hassan2018,Popel2018}, and soon refuted due to weaknesses in the evaluation methodology. Reproductions with professional translators rather than crowd workers and full documents rather than single sentences likewise concluded that HT was superior to MT in terms of both accuracy and fluency \citep{Toral2018,Laeubli2018}.


Human--machine parity claims may not hold with MT systems for broad domains such as news articles, but systems geared to narrower domains have been shown to achieve far better quality \citep[\eg,][]{Levin2017}, and it is unclear how they compare to specialised human professionals. In this paper, we propose an evaluation design that avoids the weaknesses identified in previous human--machine comparisons (\Section{Background}), and relies on metrics that are arguably better quantifiable and interpretable than adequacy and fluency judgments: error counts and edit distance (\Section{BackgroundEvaluationMT}). Evaluators are asked to flag errors in and post-edit full documents, where half of the sentences are MT and the other half are HT (\Section{Methodology}). We analyse data collected in a study involving three language pairs and ten professional translators, and find that professional translators post-edit professional HT almost as much as MT, and rate the two similarly in terms of issues with terminology, omission, and typography (\Section{Experiment}). We also contextualise our results within the ongoing discussion on human--machine parity, suggesting that further assessments will need to focus specifically on what professional translators can do better than MT systems -- and vice versa -- rather than comparing their \enquote{overall quality} (\Section{Discussion}). Our method should provide a means to assess the viability of MT in specific professional translation contexts, and may possibly help decrease resistance against the technology among professional translators.

\section{Background}
\label{sec:Background}

How to tell whether a translation is good or bad is one of the most important and one of the most difficult questions asked in connection with translation. Best practices for evaluating HT and MT differ, and assessments of human--machine parity have largely ignored the former.

\subsection{Evaluation of HT}
\label{sec:BackgroundEvaluationHT}

Quality assurance in professional translation workflows typically means manual identification of errors in (a sample of) translations. The error types depend on the quality standard. LISA, the first quality standard that gained widespread adoption in the translation industry, defines 20–123 error types and three severity levels: minor, major, and critical. SAE~J2450, originating from the automotive industry, uses fewer error types and only two severity levels: minor and major. In contrast to LISA, SAE~J2450 focusses exclusively on linguistic quality (i.e., no style and formatting, etc.). More recently, a joint academia-industry initiative has proposed the Multidimensional Quality Metrics (MQM) framework, which allows the definition of custom quality standards by choosing a subset of (weighted) error types.

The quality score of a given translation is computed as a linear combination of error counts and severity levels (i.e., weights). The error categories are defined in the quality standard; the number of errors per category and the severity of each error are determined by a single qualified rater. A translation is considered fit for purpose if its quality score does not exceed a given threshold.

\subsection{Evaluation of MT}
\label{sec:BackgroundEvaluationMT}

While there are various automatic metrics such as BLEU \citep{Papineni2002} or TER \citep{Snover2006}, human evaluation is considered the only reliable method in MT quality evaluation.\footnote{At WMT 2019, human quality judgements for the strongest MT systems were negatively correlated with BLEU, the most widely used automatic metric \citep[p.~79]{Ma2019}.} Rather than specific error categories, human evaluation of MT quality has been focussed on two rather abstract dimensions: adequacy and fluency. Human raters judge the degree to which a translation adequately expresses meaning of its source text or constitutes a fluent sentence in the target language, respectively, on either an absolute or relative scale. 5-point adjectival scales were used at the first large-scale MT evaluation campaigns, but soon replaced by relative ranking because categories such as \enquote{[the translation preserves] most meaning} and \enquote{[the translation preserves] much meaning} proved hard to distinguish \citep{KoehnMonz2006}. Relative rankings show better inter- and intra-rater agreement \citep{CallisonBurch2007}, but since they only tell if but not by how much two or more translations differ -- raters chose between better, same (tie), or worse --, the research community has lately embraced continuous Likert-like scales \cite[referred to as direct assessment, see][]{Graham2013}.

The score of a given system output, typically a few hundred to a few thousand sentences, is computed by aggregating the adequacy and fluency judgements of multiple bi- and monolingual raters, respectively. Raters are typically MT researchers \citep[\eg,][]{WMT2019} and/or crowd workers, but rarely qualified translators.

\subsection{Assessment of Human--Machine Parity}
\label{sec:BackgroundParity}

\begin{table}
    \fontsize{10.1pt}{10.1pt}\selectfont
    \renewcommand{\arraystretch}{1.5}
    \begin{tabularx}{\columnwidth}{lX}
\toprule
\textbf{Error Type} & \textbf{Definition (MQM)}                                                                                                             \\ \midrule
Terminology         & A term (domain-specific word) is translated with a term other than the one expected for the domain or otherwise specified.            \\
Omission            & Content is missing from the translation that is present in the source.                                                                \\
Typography          & Issues related to the mechanical presentation of text. This category should be used for any typographical errors other than spelling. \\ \bottomrule
\end{tabularx}
    \caption{Error types and definitions.}
    \label{tab:ErrorTypes}
\end{table}

\begin{table*}
    \renewcommand{\arraystretch}{1.3}
    \fontsize{10.1pt}{10.1pt}\selectfont
    \begin{tabularx}{\textwidth}{lXXl} \toprule
\textbf{ID} & \textbf{Source (DE)}                  & \textbf{Target (EN)}          & \textbf{Origin} \\ \midrule
1 & Dieses Arbeitspapier beschränkt sich auf die notwendigen Funktionalitäten für die Bestandsführung.&	This \textbf{work paper} is limited to the necessary functions for portfolio management. & MT \\
\midrule
2 &Das Kapitel zur Benutzerverwaltung befindet sich noch in Erstellung.&	The user administration chapter is still being prepared. & MT \\
\midrule
3 & Voraussetzungen &	Requirements\textbf{:} & HT \\
 \midrule
4 & Die in der Lohnbuchhaltung erfassten Personen müssen voll arbeitsfähig sein.&	The persons entered in payroll accounting must be fully capable of working. & HT \\
 \midrule
5 & Es werden weiters ausschliesslich Personen mit Jahreslohn adressiert und keine Personen, welche auf Stundenlohnbasis arbeiten.&	Furthermore, only persons receiving an annual salary are addressed and not persons working on an hourly wage basis. & HT \\
 \midrule
6 &Für die später beschriebenen Mutationen \textbf{inkl. Eintritt / Austritt} wird von der Web API eine Korrelations ID zurückgegeben.&	A correlation ID is returned by the Web API for the changes described later. & MT \\
 \bottomrule
\end{tabularx}
    \caption{Example of a pre-translated document in which HT and MT are interleaved, including a segment with wrong terminology (ID 1), an error in typography (3), and an omission (6). The errors in segments 3 and 6 have been fabricated for the purpose of illustration.}
    \label{tab:MaterialsExample}
\end{table*}

In summary, the evaluation of HT focusses on quality: raters are qualified translators and give feedback on specific errors (such as the number of severe terminology problems). Because qualified feedback is expensive, few segments are evaluated by a single translator. The evaluation of MT, on the contrary, focusses on quantity: many segments are evaluated by multiple raters, but those raters are not qualified and give feedback on overall quality (such as how adequate a translation is on a 100-point scale).

Given the different evaluation traditions for HT and MT, it could be assumed that a comparison of HT and MT quality would aim at combining the two. However, the first evaluation that claimed MT had reached parity with HT -- in one language pair and domain, \ie, Chinese to English news translation -- used an MT evaluation design: bilingual crowd workers rated a large number of translated sentences in terms of adequacy \citep{Hassan2018}. Two reproductions of \citeg{Hassan2018} evaluation showed that their evaluation design disadvantaged HT. Because the translated sentences were shown to raters in random order, they could not consider phenomena related to document-level cohesion, such as consistent translation of a product name throughout a news article. When raters compared full articles rather than single sentences, HT was rated significantly better than MT \citep{Laeubli2018}. Even with isolated sentences, HT was rated significantly better than MT when professional translators rather than crowd workers carried out the evaluation \citep{Toral2018}.


\section{Evaluation Design}
\label{sec:Methodology}

We propose an experimental design for combined evaluation of HT and MT that avoids the weaknesses of previous assessments on human--machine parity in translation (\Section{Background}).

\subsection{Materials}
\label{sec:MethodologyMaterials}

The evaluation is based on a source text (ST) that is segmented into either sentences or paragraphs. We obtain two translations of the entire source text: one created by a professional translator (HT), the other by the MT system (MT). The result is a segment-aligned text where each source segment (\eg, ST-1) has two translations (HT-1 and MT-1). HT is translated from scratch, \ie, without any MT system. The creator of HT has the same background as the raters (see below), but no further involvement in the experiment.

For each rater, we prepare a translation that combines ST with a mix of HT and MT. To this end, we split ST into sections of equal length. We then randomly pair each source segment with either its corresponding HT or MT, making sure to include an equal number of translations from both sources. An example is shown in \Table{MaterialsExample}. Note that the scrambling of HT and MT may introduce disfluencies, as further discussed in \Section{DiscussionExperimentalValidity}.

\subsection{Raters}

Since our evaluation involves post-editing (see below), and because translation quality is judged differently by professional translators and laypeople \citep{Toral2018}, we engage professional translators as raters. Their area of expertise matches the source text.

\subsection{Procedure}

The evaluation is organised as a task in which raters are instructed to evaluate the segments in their prepared translation (see above). Raters are told that the entire translation is MT. The primary motivation for this experimental manipulation is that we want raters to focus on evaluating segments rather than guessing if they are MT or HT. The latter would likely occur if they knew that both are present, not least because many professional translators fear \enquote{being replaced by a machine} \citep{Cadwell2018}. Translators might also be inclined to evaluate (what they believe is) MT more critically than HT because they have more negative perceptions about the former \citep{LaeubliOrregoCarmona2017}.

The evaluation of each segment involves three subtasks.
First, raters are asked to post-edit the segment. They are instructed to correct spelling and grammatical errors, but not style.
Second, raters are asked to flag the presence (but not count the number) of errors in the original target segment. We use a subset of MQM error types that has been shown to be particularly relevant for post-editing of domain-specific MT \citep{Castilho2018}, as listed in \Table{ErrorTypes}, but note that other subsets or quality standards (\Section{BackgroundEvaluationHT}) could be used instead.
Third, raters have the option to leave a comment for the segment if they wish to give more specific feedback. 

Raters complete the experiment within a fixed time frame. While the practical consideration here is limiting experimental cost, time pressure is common in professional translation \citep{EhrensbergerDow2016} and has been shown to increase cognitive function in controlled translation experiments \citep{Campbell1999}.

\subsection{Analysis}
\label{sec:MethodologyAnalysis}

\begin{table}[]
    \centering
    \renewcommand{\arraystretch}{1.3}
    \fontsize{10.1pt}{10.1pt}\selectfont
    \begin{tabular}{lrr}
\toprule
            & \multicolumn{1}{l}{HT} & \multicolumn{1}{l}{MT} \\ \midrule
Omission    & 14                     & 12                     \\
No Omission & 223                    & 226                    \\ \midrule
Total       & 237                    & 238                    \\ \bottomrule
\end{tabular}
    \caption{Contingency table for two binary variables. Raters flagged omissions in 14 segments originating from HT, and in 12 segments originating from MT. Omission does not depend on segment origin (HT vs.\ MT) according to a two-tailed Fisher's exact test ($p=0.693$). Data corresponds to \Figure{PlotsFROmission}.}
    \label{tab:ContingencyTableExample}
\end{table}

We calculate the minimum edit distance (MED) between each original and post-edited segment, as well as corpus-level HTER \citep{Snover2006} for all HT and MT segments in each target language. While HTER correlates better with human judgements of MT quality, MED is easier to interpret, particularly for individuals outside the MT research community. In reference to industry-focussed studies on post-editing \citep[\eg,][]{Volk2010}, we group post-edited segments into exact matches (MED = 0), non-exact matches (MED >0), and high effort (MED >5).

Besides descriptive statistics, we test if the presence of errors and post-editing effort depends on whether target segments originate from HT or MT. Target segment origin is our binary independent variable, and we test if its proportion varies among the proportion of a single binary dependent variable using a two-tailed Fisher's exact test as implemented in \textit{R} \citep{Bailey1995}. An example is shown in \Table{ContingencyTableExample}.

\section{Experimental Results}
\label{sec:Experiment}

We use the evaluation design described in the previous section to compare HT to MT in an experiment with three language pairs and ten professional translators. The study is conducted within the language services department of a multinational insurance company.

\subsection{MT System}
\label{sec:ExperimentMTSystem}

\begin{table}
    \centering
    \renewcommand{\arraystretch}{1.3}
    \fontsize{10.1pt}{10.1pt}\selectfont
    \begin{tabular}{lrrr} \toprule
              & \multicolumn{1}{l}{DE--EN} & \multicolumn{1}{l}{DE--FR} & \multicolumn{1}{l}{DE--IT} \\ \midrule
\multicolumn{1}{l}{\textbf{Segments}} \\
~~~ID            & 527,526                    & 1,177,704                  & 905,302                    \\
~~~OOD           & 20,000,000                 & 7,760,035                  & 6,925,296                  \\
~~~Ratio~~~      & 10:1                       & 6:1                        & 7:1                        \\[0.5em]
\textbf{Terms}         & \multicolumn{1}{l}{}       & \multicolumn{1}{l}{}       & \multicolumn{1}{l}{}       \\
~~~Train         & 10,332                     & 11,551                     & 10,537                     \\
~~~Test          & 3,256                      & 4,915                      & 4,817                      \\ \bottomrule
\end{tabular}
    \caption{Training Data}
    \label{tab:TrainingData}
\end{table}

\begin{table*}
        \renewcommand{\arraystretch}{1.4}
        \fontsize{10.1pt}{10.1pt}\selectfont
        \centering
        \begin{adjustwidth}{-2mm}{}
        \begin{tabular}{lrrrrrrrrrrrr} \specialrule{1pt}{0.5em}{0.5em}
                        & \multicolumn{4}{c}{\textbf{DE--EN}}                                                                                             & \multicolumn{4}{c}{\textbf{DE--FR}}                                                                                             & \multicolumn{4}{c}{\textbf{DE--IT}}                                                               \\[0.2em]
                        & \multicolumn{2}{l}{\textbf{HT} (N=150)}                                 & \multicolumn{2}{l}{\textbf{MT} (N=150)}                                 & \multicolumn{2}{l}{\textbf{HT} (N=237)}                                 & \multicolumn{2}{l}{\textbf{MT} (N=238)}                                 & \multicolumn{2}{l}{\textbf{HT} (N=244)}                  & \multicolumn{2}{l}{\textbf{MT} (N=248)}                  \\ [0.5em] \specialrule{0.5pt}{0pt}{0.5em}
\textbf{Error Analysis} &                            &                                   &                            &                                   &                            &                                   &                            &                                   &                          &                      &                          &                      \\
~~~Terminology          & 8                          & \hspace{-3mm}(5.33)                          & 15                         & \hspace{-3mm}(10.00)                         & 27                         & \hspace{-3mm}(11.39)                         & 39                         & \hspace{-3mm}(16.39)                         & 18                       & \hspace{-3mm}(7.38)             & 19                       & \hspace{-3mm}(7.66)             \\
~~~Omission             & 1                          & \hspace{-3mm}(0.67)                          & 5                          & \hspace{-3mm}(3.33)                          & 14                         & \hspace{-3mm}(5.91)                          & 12                         & \hspace{-3mm}(5.04)                          & 4                        & \hspace{-3mm}(1.64)             & 1                        & \hspace{-3mm}(0.40)             \\
~~~Typography           & 3                          & \hspace{-3mm}(2.00)                          & 4                          & \hspace{-3mm}(2.67)                          & 5                          & \hspace{-3mm}(2.11)                          & 3                          & \hspace{-3mm}(1.26)                          & 8                        & \hspace{-3mm}(3.28)             & 6                        & \hspace{-3mm}(2.42)             \\[0.5em]
\multicolumn{13}{l}{\textbf{MED}}                                                                                                                                                                                                                                                                                                                                                               \\
~~~\textgreater 0       & *\,20 & \hspace{-3mm}(13.33) & *\,37 & \hspace{-3mm}(24.67) & *\,67 & \hspace{-3mm}(28.27) & *\,90 & \hspace{-3mm}(37.82) & 65                       & \hspace{-3mm}(26.64)            & 50                       & \hspace{-3mm}(20.16)            \\
~~~\textgreater 5       & 12                         & \hspace{-3mm}(8.00)                          & 19                         & \hspace{-3mm}(12.67)                         & *\,53 & \hspace{-3mm}(22.36) & *\,75 & \hspace{-3mm}(31.51) & 30                       & \hspace{-3mm}(12.30)            & 27                       & \hspace{-3mm}(10.89)            \\[0.5em]
~~~min                  & 0                          &                                   & 0                          &                                   & 0                          &                                   & 0                          &                                   & 0                        &                      & 0                        &                      \\
~~~max                  & 85                         &                                   & 43                         &                                   & 118                        &                                   & 150                        &                                   & 34                       &                      & 130                      &                      \\
~~~avg                  & 1.56                       &                                   & 2.89                       &                                   & 6.89                       &                                   & 7.83                       &                                   & 2.39                     &                      & 2.92                     &                      \\
~~~med                  & 0                          &                                   & 0                          &                                   & 0                          &                                   & 0                          &                                   & 0                        &                      & 0                        &                      \\
~~~sd                   & 7.85                       &                                   & 7.86                       &                                   & 17.41                      &                                   & 17.43                      &                                   & 6.17                     &                      & 13.07                    &                      \\[0.5em]
\textbf{HTER}           &                            &                                   &                            &                                   &                            &                                   &                            &                                   &                          &                      &                          &                      \\
~~~Corpus-level         & \multicolumn{1}{l}{2.22}   & \multicolumn{1}{l}{}              & \multicolumn{1}{l}{4.71}   & \multicolumn{1}{l}{}              & \multicolumn{1}{l}{7.42}   & \multicolumn{1}{l}{}              & \multicolumn{1}{l}{7.99}   & \multicolumn{1}{l}{}              & \multicolumn{1}{l}{3.67} & \multicolumn{1}{l}{} & \multicolumn{1}{l}{3.81} & \multicolumn{1}{l}{} \\ \bottomrule
\end{tabular}
        \end{adjustwidth}
        \caption{Results. Counts denote the number of segments for which a given variable holds true for HT or MT, respectively; relative numbers are shown in brackets. Pairs of significantly different proportions according to a two-tailed Fisher's exact test (at $p\leq0.05$) are marked with *. Example: In DE--FR, 5/237 HT segments and 3/238 MT segments contain a typographical error. The difference is not statistically significant. Visualisations and $p$ values are shown in Figures~\ref{fig:PlotsEN}--\ref{fig:PlotsIT}.}
        \label{tab:Results}
\end{table*}

We train a Transformer (big) model \citep{Vaswani2017} as implemented in Sockeye \citep{Sockeye} with FFN size 2048 for each language pair. The training data is listed in \Table{TrainingData}. We combine publicly available out-of-domain data (OOD) from OPUS \citep{OPUS}, from which we discard the lowest-scoring 75\percent by means of dual conditional cross-entropy filtering \citep{JunczysDowmunt2018}, with in-domain data (ID). We oversample ID to match OOD where possible, with a maximum oversampling factor of 10.

We also integrate domain-specific terminology by means of data augmentation \citep{Dinu2019}. We use two different sets of terms for training and testing (\ie, use in production). For training, we automatically filter the insurance company's full terminology, removing terms with low frequencies in the training data for reasons of time efficiency, and using a stop word list to remove terms that occur frequently in regular text (\enquote{normal words}). In addition, we discard terms in 30\percent of the training segments to increase robustness in constraint-free scenarios. For testing, we use a smaller terminology that was narrowed down by the company's professional terminologists.

\subsection{Texts and Raters}

For each language pair, we select a document that contains terminology and language specific to the company's insurance sector: the description of business processes in a customer application (DE--EN) and a text on specialist training in sales (DE--FR, DE--IT).

We have all three documents translated by external translators who are regularly contracted by the company. We also translate the documents using the MT systems described above, and prepare a pre-translated version of each document in which half the target segments stem from the external translators and the other half from the MT system (\Section{MethodologyMaterials}).

The raters participating in the experiment are in-house translators at the company, and have not previously seen these documents. The number of raters differs between language pairs: four raters each for DE--FR and DE--IT, and two for DE--EN. Each rater is allocated 150 consecutive segments of the document, so the number of experimental items (segments) amounts to 600 for DE--FR and DE--IT, and to 300 for DE--EN.

The raters were given 90 minutes to complete the task. Two raters for DE--FR and one rater for DE--IT did not finish in time, reducing the number of items in our analysis to 475 and 492, respectively.

\subsection{Error Analysis}
\label{sec:ResultsErrorAnalysis}

Experimental results are listed in \Table{Results}. We first analyse the proportion of segments that contain at least one terminology, omission, or typography error originating from HT and MT. 
The number of segments with terminology errors is higher for MT than HT. While almost twice as many segments are affected in DE--EN, the difference is less marked in DE--FR, and very small in DE--IT.
Omissions are found in more segments originating from MT in DE--EN, and in more segments originating from HT in DE--FR and DE--IT. The number of segments containing omissions are considerably lower in DE--EN and DE--IT than in DE--FR.
In terms of typography, the number of affected segments is low for both HT and MT. HT is slightly better than MT in DE--EN, and slightly worse in DE--FR and DE--IT.

The proportion of erroneous segments is similar for HT and MT overall. A two-tailed Fisher's exact test shows no significant difference between HT and MT in any error category and language pair. $p$-values are shown in Figures~\ref{fig:PlotsEN}--\ref{fig:PlotsIT}.




\subsection{Post-editing Effort}
\label{sec:ResultsEditDistance}

We compute corpus-level HTER for all HT and MT segments in each language pair (\Table{Results}, last row). We observe very low scores overall, and small differences between HT and MT in DE--FR and DE--IT.

We also compute MED between each pre-translated and post-edited target segment. Descriptive statistics are listed in \Table{Results}. In all language pairs, raters post-edited less characters in HT on average (avg), but again, the differences are small, particularly for DE--IT. The segment that required most post-editing (max) stemmed from HT in DE--EN, and from MT in DE--FR and DE--IT.

We observe a low number of segments that required any post-editing at all. The proportion of these segments is referred to as >0 in \Table{Results}. For example, only 37 out of 150 MT segments in DE--EN were post-edited; raters decided that raw MT was good enough for the remaining segments. However, the proportion of segments that needed any editing was even lower for HT in DE--EN, significantly so according to a two-tailed Fisher's exact test ($p$$\leq$.05). The difference between the proportion of segments with an MED of more than five characters (>5), on the other hand, is not significant ($p$=0.255) in DE--EN. In DE--FR, both >0 and >5 segments are significantly more frequent in MT (both at $p$$\leq$.05). In DE--IT, where raters post-edited more HT than MT segments (see >0), the difference is not significant at $p$=0.110 and $p$=0.674, respectively.



\section{Discussion}
\label{sec:Discussion}

We discuss design decisions in our evaluation and alternative approaches to inference testing, and contextualise our results within the ongoing discussion on human--machine parity in language translation.

\subsection{Experimental Validity} 
\label{sec:DiscussionExperimentalValidity}

Our evaluation is based on pre-translated documents in which target segments from HT and MT are interleaved (\Table{MaterialsExample}). In contrast to other MT quality evaluation experiments \citep[\eg,][]{Green2013,Hassan2018}, this enables raters to consider document-level context, but the shuffling of MT and HT may introduce disfluencies that would not occur if all segments stemmed from either MT or -- particularly -- HT. In DE--FR, for example, the German term \textit{Einzelfirma} (sole proprietorship), which occurred in seven source segments, was translated as \textit{raison individuelle} and \textit{entreprise individuelle} by HT and MT, respectively. The first three instances were translated by MT, and noting the inconsistency with the fourth instance translated by HT, the rater in charge flagged the segment as erroneous and commented that \enquote{[the term translations] should be harmonised}. The MT system's translation was consistent with the company's terminology database (TB) in this case, and the flagging of HT as erroneous was correct. However, if MT and HT used different translations for a term not specified in the TB, the translation introduced second would likely be marked as wrong even if it was used consistently within HT and MT. This may increase the number of terminology errors overall, but since the order in which MT and HT appear in documents is randomised in our evaluation design, it would not disadvantage one over the other with sufficient sample size. We also note that combining segments from different sources is common in professional translation workflows: when translations for adjacent source segments are retrieved from a translation memory (TM), these translations may (and typically will) stem from different documents and translators. The documents we prepared for our experiment are what translators would normally see in their computer-aided translation (CAT) tool, with HT corresponding to exact matches, except that segment origin (HT or MT) is not shown in the experiment.

We did not use a CAT tool in our experiment, but presented the pre-translated documents as spreadsheets with dedicated columns for error annotations and comments. A downside of this design decision is that the company's TB was not directly integrated into the translation environment. In the CAT tool that the in-house translators (the raters in this experiment) use in their daily work, terms contained in the TB are highlighted in source segments, and term translations are shown in a dedicated window. While raters had access to the TB during the experiment, it is likely that they missed a few terminology errors because terms were not highlighted in the experiment. On the contrary, we noticed that they marked a variety of other mistakes as terminology errors, such as wrong choice of pronoun (\eg, \textit{que} instead of \textit{soi} in DE--FR) or wrong verb forms (\eg, \textit{data already exists} instead of \textit{data already exist} in DE--EN). Since raters blindly evaluated HT and MT segments the same way, this may affect the true number of terminology errors in our analysis, but not the proportion between errors in HT and MT.

The blind evaluation of pre-translated segments -- the fact that we did not tell raters that half of the pre-translations were HT, and that we did not show that pre-translations originated from different sources (HT and MT) -- is another design decision that warrants discussion. Whether a pre-translated segment was retrieved from a TM (as an exact or fuzzy match) or an MT system is important information to professional translators and thus prominently shown in CAT tools. However, beliefs about (non-)presence of MT have been shown to impact how willing people are to tolerate translation mistakes \citep{Gao2014}, and surveys have shown that professional translators tend to have negative perceptions about MT \citep{LaeubliOrregoCarmona2017,Cadwell2018}. Our experimental manipulation was aimed at fostering equal rigour in evaluating HT and MT, and preventing raters from guessing if segments are HT or MT rather than focussing on actual evaluation.

\subsection{Statistical Analysis}
\label{sec:StatisticalAnalysis}

A limitation of using contingency tables (see \Table{ContingencyTableExample} for an example) is that we can only use categorical variables as dependent variables. To that end, we binarised MED with fixed and arguably arbitrary thresholds (>0 and >5; see \Section{MethodologyAnalysis}). Predicting MED in a regression model would seem more appropriate, and offers the advantage of accommodating further predictors such as segment length, but violated the assumption of normally distributed residuals in our data even when extreme values were removed. Futher analysis, including factors other than origin (HT/MT) that may explain the variance in presence of errors and post-editing distance, is left to future work.

We use Fisher's exact test to analyse contingency tables, the null hypothesis being that the likelihood of a segment showing a certain property -- such as containing wrong terminology or having been post-edited (MED >0) -- is not influenced by its origin (HT or MT). Fisher's exact test has been criticised as rather conservative \citep[see][]{AndresTejedor1995}, but is more appropriate than $\chi^2$ or $G$ tests of independence when sample sizes are small \citep{RuxtonNeuhaeuser2010}.\footnote{Using a $\chi^2$ or $G$ test of independence has no effect on any finding of (non-)significance reported in this paper. We observe the largest difference when testing for independence of origin and omission in DE--EN with a $G$ test ($p$=0.085) instead of a two-tailed Fisher's exact test ($p$=0.214, see \Figure{PlotsENOmission}).}

It would also be desirable to include more raters in the experiment. The limited number of participants is often criticised in translation experiments, justifiably so because translation performance varies considerably between individuals \citep[\eg,][]{KoehnGermann2014}. With sufficient participants, this variance can be accounted for by means of mixed-effects modelling \citep{Green2013}, but quite apart from budgetary constraints, there may just not be enough qualified raters in domain-specific settings. The in-house translation department we work with in this study, for example, employs 2--4 specialised translators per language pair. Non-experts who could be involved to increase the number of raters have been shown to evaluate MT less critically \citep{Toral2018}. In the present study, we prioritised rater qualification over quantity.

\subsection{Human--Machine Parity?}

Our results illustrate that the question whether MT quality reaches parity with HT is a matter of definition. \citet{Hassan2018}, who analysed quality judgements by crowd workers in Chinese to English news translation, concluded that parity was reached because the difference between judgements of HT and MT is not statistically significant. The same holds for our experiment: professional translators flagged errors in segments originating from HT and MT, and the proportion of erroneous HT and MT segments does not differ significantly for any error type and language pair (\Section{ResultsErrorAnalysis}). This is mainly because error rates are fairly low for both HT and MT, which indicates that both translation methods achieve high quality. However, MT produced more erroneous segments than professional translators (HT) overall, and the fact that statistical tests (\Section{StatisticalAnalysis}) find no significant difference between HT and MT either means that there really is none, which would imply parity, or that the number of analysed segments (the sample size) is too small to infer a significant difference. Consider the proportion of segments with omissions in DE--EN (\Table{Results}): 1/150 in HT vs.\ 5/150 in MT. Omissions are rare in both, and the difference is attributed to chance ($p$=0.214, see also \Figure{PlotsENOmission}), but in the very document we analysed, omissions were five times more common in MT segments nonetheless. If assessing human--machine parity was the aim of our study, a larger sample size would be imperative to come to understand if such effects are true or random. Nevertheless, the observation that MT produced less erroneous segments than HT in at least one language pair per error type in our experiment -- except for terminology, where MT only came close to HT in DE--IT with 19/248 vs.\ 18/244 erroneous segments, respectively -- is noteworthy.


While our error analysis was limited to three specific phenomena -- terminology, omission, and typography -- the comparison of pre-translated to post-edited segments yields insights about HT and MT quality overall. MT produced significantly more segments that needed post-editing at all (MED >0) in DE--EN and DE--FR. In DE--EN, however, the proportion of segments that needed substantial post-editing (more than five characters, \ie, MED >5) was not significantly higher in MT, and in DE--IT, the number of segments that needed any (MED >0) and substantial (MED >5) post-editing was lower in MT than in HT. This is a remarkable finding, given that HT was produced by an expert translator with experience in the textual domain we investigate. The implication here is that domain-specific MT (\Section{ExperimentMTSystem}) achieves strong results, and it may be insightful to contrast it with generic MT. Moreover, feedback from raters, who had the option to leave a comment for each segment, does not suggest that the experimental manipulation -- the mixture of MT with HT -- was noticeable. In one particular instance, a rater commented \enquote{NMT hat überkorrigiert} (\enquote{NMT has overcorrected}), when in fact the segment in question originated from HT.

\section{Conclusion}
\label{sec:Conclusion}

In a blind evaluation, ten specialised translators post-edited and flagged errors in pre-translated documents in which domain-specific MT was interleaved with professional HT. The evaluation comprised three language pairs: DE--EN, DE--FR, and DE--IT. MT required more post-editing than HT on average, but surprisingly, the difference is not significant in DE--IT, where MT produced more segments that needed no post-editing at all, and slightly less segments that needed substantial post-editing. We also analysed if the proportion of segments that contain wrong terminology, omissions, or typographical errors varies among HT and MT, and found no significant dependency in any language pair. MT produced considerably more segments with wrong terminology in two out of three language pairs, but slightly less segments with omissions or typographical errors in at least one language pair each.

Apart from implying that MT can now reach remarkable quality in domain-specific settings, our results show that professional translators may post-edit professional HT almost as much as MT, and tend to rate the two similarly in terms of issues with terminology, omission, and typography. The caveat here and an aspect that warrants further investigation is that we made our participants believe that the HT they were evaluating was MT. From a methodological point of view, it would be interesting to test if this experimental manipulation would also work the other way around, and analyse if translators treat HT and MT differently depending on what they believe it is. From a more practical perspective, it might also be worth exploring whether the proposed evaluation design could help demonstrate the potential benefits of MT to people who are still sceptical about the technology.

\bibliographystyle{acl}
\fontsize{10.1pt}{10.1pt}\selectfont
\bibliography{references}

\appendix

\section{Visualisations}

\begin{figure*}[h]
    \centering
    \input{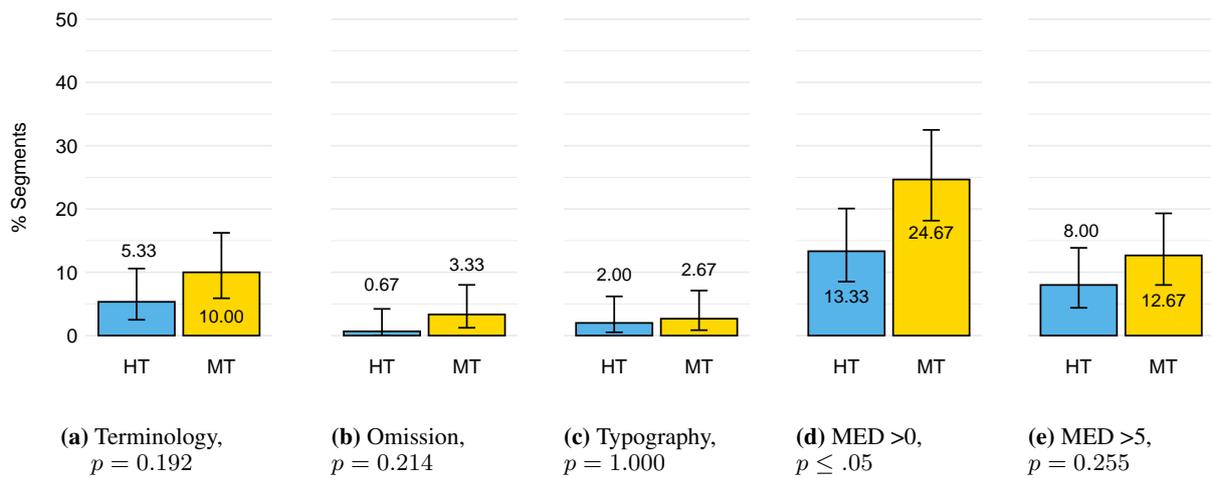}
    \caption{German--English}
    \label{fig:PlotsEN}
\end{figure*}

\begin{figure*}
    \centering
    \input{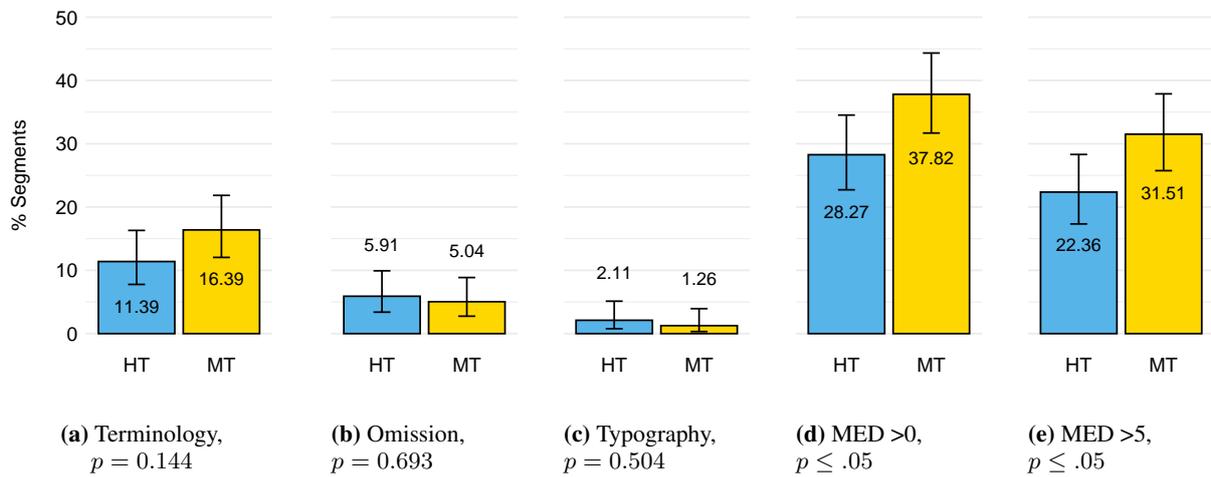}
    \caption{German--French}
    \label{fig:PlotsFR}
\end{figure*}

\begin{figure*}
    \centering
    \input{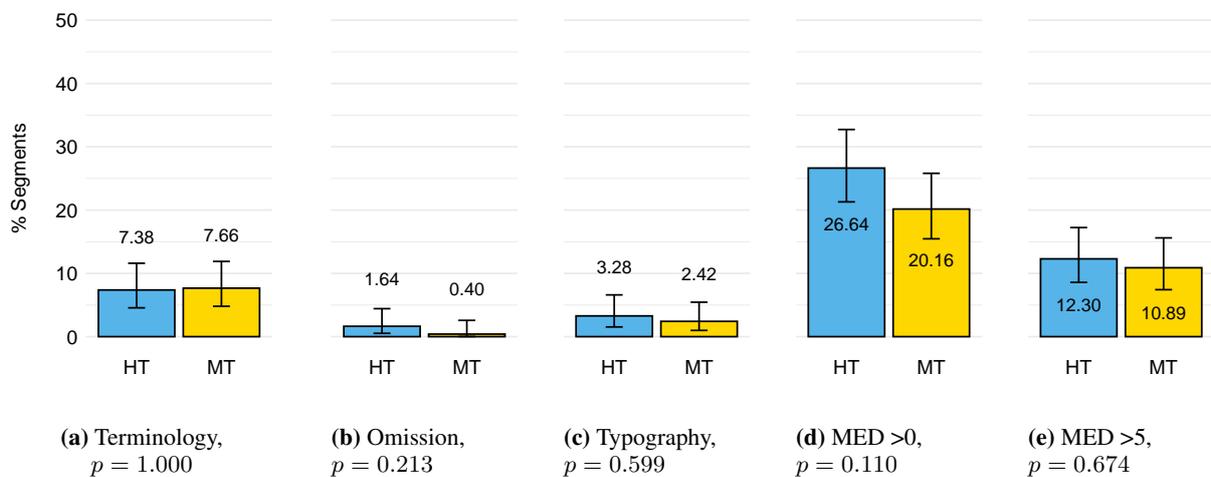}
    \caption{German--Italian}
    \label{fig:PlotsIT}
\end{figure*}

Figures~\ref{fig:PlotsEN}--\ref{fig:PlotsIT} visualise the main results listed in \Table{Results}. Each plot corresponds to a 2x2 contingency table for two binary variables (\Table{ContingencyTableExample}). We compute $p$-values using Fisher's exact test (see \Section{MethodologyAnalysis}). Error bars show 95\percent confidence intervals.

\end{document}